\documentclass{article}
\usepackage{amsmath,amssymb,graphicx,url}
\usepackage{amssymb,amsfonts}
\usepackage{amsmath,amssymb,graphicx,url}
\usepackage{natbib}

\newtheorem{theorem}{Theorem}
\newtheorem{proposition}{Proposition}
\newtheorem{lemma}{Lemma}
\newtheorem{corollary}{Corollary}

\def\x{{\bf x}}
\def\y{{\bf y}}

\def\p{{\bf p}}
\def\q{{\bf q}}

\def\w{{\bf w}}
\def\l{{\bf l}}
\def\0{{\bf 0}}

\def\a{{\bf a}}

\def\f{{\bf f}}

\def\subst{{\rm Subst}}

\def\0{{\bf 0}}

\title{Prediction of Locally Stationary Data Using Expert Advice}

\author{Vladimir V'yugin, Vladimir Trunov\\
  Institute for Information Transmission Problems,\\
  Russian Academy of Sciences}

\date{}

\begin{document}

\maketitle

\begin{abstract}

The problem of continuous machine learning is studied.
Within the framework of the game-theoretic approach, when for calculating the next forecast,
no assumptions about the stochastic nature of the source that generates the data flow are used --
the source can be analog, algorithmic or probabilistic, its parameters can change at random times,
when building a prognostic model, only structural assumptions are used
about the nature of data generation. An online forecasting algorithm
for a locally stationary time series is presented. An estimate of the efficiency
of the proposed algorithm is obtained.
\end{abstract}


\section{Introduction}

Predicting data coming from a ``black box'' is one of the main tasks of machine learning.
In this case, no stochastic assumptions about data source is used. The data comes online as a time series
consisting of pairs of the form (``signal'', ``response''). The data source can be an analog,
deterministic (algorithmic) or stochastic process. In this case, we will use
simple structural assumptions about the source of the data.

In this paper, an approach is proposed in which training is performed on small subsamples of the main sample,
forecasts of the constructed predictive models are combined into one common forecast based on the known aggregation methods.
The general scheme of the online learning process is as follows. The learning process occurs at discrete times
in steps $t=1,2,\dots $. At the next step $t$, according to the data from the subsample,
from the data observed in the past, a local predictive model (expert predictive strategy) is defined to obtain
a response to the signal. As a rule, this is a regression function built on the observed segment of the time series.
Thus, at step $t$, there are $t$ predictive models built from the corresponding subsamples from the past.
After that, the signal $\x_t$ is observed and all the expert predictive strategies
built on steps $1,2,\dots,t$ present their response predictions.
Predictor builds its response prediction by aggregating the experts predictions.

The problem of online aggregation of forecasts is solved within the framework of the theory
of predictions using Prediction with Expert Advice. This approach is
widely represented in the scientific literature on machine learning(see \citealt{Vov1998},
\citealt{cesa-bianchi}, \citealt{Vyu2022}).

After the predictions are presented, the source (the corresponding generator) produces the true response $y_t$, and
the experts and Predictor calculate their losses due to the difference between their predictions and the response.

In mathematical statistics, when building predictive models, one often use
stochastic assumptions about the nature of the data. In this work, to build predictive models
online methods of machine learning are used within the framework of the game-theoretic approach,
while stochastic data models are not used.

When constructing predictive strategies, assumptions about the structure of the data generation method can be used.
The following data generation scheme is assumed that there are several generators,
which, replacing each other, generate a time series, which, thus, is divided into subsamples -- areas of stationarity.
The device of the generators is unknown to the experts and Predictor. Each area of stationarity can be studied by machine
learning methods based on the results of the generator, i.e., according to the data from the stationarity region,
the corresponding local predictive algorithm(local predictive model) will be built tied to a generator
that can be applied to other stationarity domains generated by the same generator.

In the theory of prediction with expert advice, the efficiency of an aggregating algorithm is evaluated
using the concept of a regret, which is the difference between the total (cumulative) losses of the aggregating
algorithm and the total losses of the expert algorithm accumulated over the entire prediction period.
The goal of the aggregating algorithm is to minimize the regret with respect to each expert
strategy (see ~\citealt{Vyu2022}, \citealt{cesa-bianchi}, \citealt{Vov1998}).

In another, more general, formulation of the forecasting problem, the regret
of the aggregating algorithm with respect to arbitrary sequences of expert
strategies is minimized:
a series of steps at which predictions are made is divided into segments.
Each segment is assigned its own expert; the sequence of segments
and corresponding experts is called a composite expert. The purpose of
the algorithm changes -- now it must predict in such a way that it is not
worse than each composite expert.
Accordingly, the concept of the algorithm regret is modified --
now it is the difference between the total loss of the algorithm and the total loss of
the sequence of experts.
This change allows us to more accurately simulate real life conditions,
when the nature of outcomes can change over time and different experts can predict
with varying degrees of success depending on the current trend. the corresponding
the algorithm is called Fixed Share~\citet{HeW98}.
In the work~\citet{BoW1998} a further generalization of the Fixed Share method
was proposed -- the method of mixing past posterior distributions MPP (Mixing Past Posteriors).
The cumulative loss of the aggregation algorithm are related to the loss of any convex combinations
of the  experts. The concept of regret also changes. Now the total loss of the algorithm
is compared with the total loss of convex combinations of expert strategies
(see details in \citet{Vyu2022} and \citealt{BoW1998}).
In this work, we apply this approach to construct an algorithm for predicting locally stationary data.

A characteristic feature of the problem considered in this work is the absence
of a predetermined set of competing expert strategies, as was the case in the
works cited above. Instead, new expert strategies are being built at every step of
the online learning process. The predictor must aggregate at each step the forecasts
of all the expert strategies built by that time.

Let us briefly describe the proposed approach.
Expert strategies (local predictive models) are automatically built up online depending on
the observed real data. At each step, a new expert predictive strategy is introduced
that reflects the local properties of the observed part of the time series (subsample).
Forecasts of all predictive strategies built up to this point are combined
to the Predictor forecast using one of the aggregation methods.

The general scheme of learning with a teacher using expert strategies has the form of
a game with participants: Predictor, experts $i\in{\cal N}$. At each step $t$ of the game,
each expert $i$ observes the signal $\x_{t}$ and
provides its prediction $f_{i,t}=f_{i,t}(\x_t)$, Predictor calculates its prediction
$\gamma_t$. After that, the the true response $y_t$ is presented and the losses $l_{i,t}=\lambda(f_{i,t},y_t)$
of the experts and the loss $h_t=\lambda(\gamma_t,y_t)$ of Predictor are calculated,
where $\lambda(\gamma,y)$ is a loss function that takes non-negative values.

We assume that the data source has several response generation modes, so the responses generated by it
the data is broken down into appropriate time ``stationarity intervals''.
Each stationarity interval corresponds to a certain data source operation mode and is characterized
by a valid predictive model -- an expert strategy whose parameters are determined by
the ``stationarity interval''.

The parameters of the valid prognostic model corresponding to a given mode of operation of the source can be limited
at the first appearance of stationarity intervals (corresponding to the operating mode of the source).
The assumption is used that after determining its parameters, the predictive model
has the property of validity on other intervals of stationarity corresponding to the
same mode of operation of the source (generator).

Since the boundaries of the stationarity intervals are unknown to Predictor,
the expert predictive models are built at each training step. Some of these models are valid, i.e.,
they be trained on data generated by some generator, the rest will be
not valid, i.e.they do correspond to data from any stationarity interval. Thus,
at each stage of forecasting, we have a collection of valid and invalid local predictive models,
from which we compose a single effective predictive model of Predictor.
The constructed predictive models compete with each other at every moment of time,
so we will combine (aggregate) them using using methods of the theory of prediction with expert advice.
The main result of this work is the construction and study of an algorithm for predicting
a locally stationary time series, which aggregates all the constructed predictive models,
highlighting the forecasts of valid local predictive models.

The proposed approach is implemented in the form of the ${\bf GMPP}$ algorithm, and the theoretical
bound of the loss of this algorithm will be obtained. 
\section{Preliminaries}

\subsection{Prediction with expert advice}

Let $\lambda(\gamma,y)$ be the a loss function, where $\gamma$ is a forecast, $y$ is an outcome (response).
The loss function accepts non-negative real numbers as values. The simplest example of a loss function
for a regression problem: in the case when the outcomes and forecasts are real numbers from $\cal{R}$,
the square loss function $\lambda(\gamma,y)=(\gamma-y)^2$ is used.

The general scheme of learning with a teacher using expert strategies is given below in the form of a game
with participants: Predictor, experts $i\in{\cal N}$. At each step $t$ of the game, each Expert $i$ observes
the signal $\x {t}$ and provides its prediction $f_{i,t}=f_{i,t}(\x_t)$, Predictor provides
its prediction $\gamma_t$., after that the true response is revealed and experts calculate theirs losses
$l_{i,t}=\lambda(f_{i,t},y_t)$, Predictor calculates its loss $h_t=\lambda(\gamma_t,y_t)$,
where $\lambda(\gamma,y)$ is a loss function that takes non-negative values.

The specificity of the problem lies in the fact that the number of experts is not limited --
each expert $i$, or rather,
the prediction function $f_{i,t}=f_{i,t}(\x_t)$, will be built at step $i$ and used in subsequent steps.
Therefore, we have to assume in advance that the number of experts is
infinite and consider the problem of prediction using forecasts of expert strategies for an infinite number of experts.

Let us present the classical formulation of the prediction problem using expert forecasts for the case when
the number of experts is infinite.
We assume that there is an infinite set of expert strategies $i\in{\cal N}$, where ${\cal N}$
is the set of all nonnegative integer numbers (or an initial segment of this set).\footnote{The second case
is the classical setting considered in \citet{Vov1990}, \citet{Vov1998}, in which the algorithm is
trained using expert forecasts from a predetermined finite set, in this case $\cal{N}$ is the
initial segment of the natural series.}

The order of actions of players and access to information is determined by the following online protocol.

\noindent{\bf Protocol 1}

{\small
\medskip\hrule\hrule\medskip

\medskip
\noindent{\bf FOR} $t=1,\dots ,T$
\begin{enumerate}

\item
Each expert $i\in\cal{N}$ presents its own prediction $f_{i,t}$.
\item
 Predictor presents its prediction $\gamma_t$.
\item Get the outcome $y_t$ and calculate the loss of each Expert $i$:
$l_{i,t}=\lambda(f_{i,t},y_t)$ and the loss
$h_t=\lambda(\gamma_t,y_t) $of Predictor.
\end{enumerate}

\noindent \hspace{2mm}{\bf ENDFOR}
\medskip\hrule\hrule\medskip
}
\medskip

Total loss $L_{i,T}$ of an arbitrary expert $i$
and the total loss $H_T$ incurred by Predictor in the first $T$ steps are defined as
$L_{i,T} = \sum\limits_{t=1}^T l_{i,t}$ and $H_T = \sum\limits_{t=1}^T h_t$, respectively.

Experts can get their predictions in one way or another, which does not matter in this game.
The predictor must have his own strategy for calculating $\gamma_t$ predictions.
The construction of such a strategy is the main task in the construction of a forecasting method.
The predictor can use all the information that is known for his move, in particular, he can use
the current and past predictions of experts, past outcomes, as well as losses of experts in
the past steps of the game.
The efficiency of Predictor relative to the expert $i$ is measured by the
regret $R_{i,T}=H_T-L_{i,T}$.
The task of Predictor is to minimize the regret in relation to each of the experts.

The Predictor's strategy is based on the use of weights assigned to experts depending on their
losses in the past. First, the initial values of weights $w_{i,1}$ at $i\in\cal{N}$ are set.
For example, $w_{i,1}=\frac{2}{c(i+1)\ln^2(i+1)}$, where
$c=\sum_{i\in {\cal N}}\frac {1}{(i+1)\ln^2(i+1)}$.\footnote{ $\frac{1}{\ln3}<c<\frac{1}{\ln3}$.
As $w_{i,1}$, elements of any convergent series are suitable. }

The Predictor's strategy is based on the use of weights, which are assigned to experts
depending on their losses in the past.

First, the initial values of the weights $w_{i,1}$ at $i\in\cal{N}$ are specified. For example, $w_{i,1}=\frac{1}{c(i+1)\ln^2(i+1)}$,
where $c=\sum_{i\in {\cal N}} \frac{1}{(i+1)\ln^2(i+1)}$.\footnote{
{{$\frac{1}{\ln3}<c<\frac{1}{\ln3}$, numerically $c\approx 2.10974$.
As $w_{i,1}$ elements of any convergent series are suitable.}
}
}
At the end of each step $t$, we update the weights using the exponential weighting method:
\begin{eqnarray}
w_{i,t+1}=w_{i,t}e^{-\eta l_{i,t}}
\label{weight-1}
\end{eqnarray}
for each $i\in\cal{N}$, where $\eta>0$ is a learning parameter

 Weights are normalized:
$$
w^*_{i,t}=\frac{w_{i,t}}
{\sum\limits_{j\in\cal{N}}w_{j,t}}.
$$
(see~\citealt{Vyu2022},~\citealt{Vov1990},~\citealt{cesa-bianchi} for details).

The quantity
$$
m_t=-\frac{1}{\eta}\sum\limits_{i\in\cal{N}} w^*_{i,t}e^{-\eta\lambda(f_{i,t} ,y)}
$$
is called the exponentially mixed loss (mixloss) and the quantity $M_T=\sum_{t=1}^T m_t$ is called
the cumulative (total) exponentially mixed loss at steps $t=1,\dots ,T$.\footnote{In
statistical physics the quantity $m_t$ is called the statistical sum. It is easy to see that these
 are finite.} $L_{i,T}=\sum_{t=1}^T l_{i,t}$ -- total loss of
th expert $i$ for the first $T$ steps. These values underlie the analysis of predictive algorithms.
\begin{proposition}\label{prop-1s} For any expert $i$
\begin{eqnarray*}
M_T\le L_{i,T}+\frac{1}{\eta}\ln\frac{1}{w_{i,1}}
\end{eqnarray*}
for every $T$.
A typical bound for $M_T$ is presented below.
\end{proposition}

\emph{Proof} Let $\w^*_t=(w^*_{i,t}:i\in\cal{N})$ be the normalized weights and
$\f_t=(f_{i,t}:i\in\cal{N})$ be the experts forecasts at step $t$.
Predictor's forecast is denoted by $f_t$. It follows from the definition that
\begin{eqnarray}
m_t=-\frac{1}{\eta}\sum_{i\in\cal{N}}
e^{-\eta\lambda(f_{i,t},y_t)}w^*_{i,t}=-\frac{1}{\eta}\ln\frac{W_{t+1} }{W_t}
\label{h-t-bound-1}
\end{eqnarray}
for all $t$, where $W_t=\sum\limits_{i\in\cal{N}} w_{i,t}$ and $W_1=1$.
From (\ref{weight-1}) we have $ w_{i,T+1}=w_{i,1}e^{-\eta L_{i,T}}$.
By telescoping, we obtain for any expert $i$ a time-independent bound
$M_T=-\frac{1}{\eta}\ln W_{T+1}\le L_{i,T}+\frac{1} {\eta}\ln\frac{1}{w_{i,1}}$ for all $T$.
$\Box$

The method of calculating Predictor's forecast is specified in Section~\ref{AA}.

\subsection{MPP and Fixed Share methods}\label{FS-1}

In what follows, we will use an important generalization of the classical prediction scheme
using expert strategies --
the method of mixing past posterior distributions of experts -- MPP.

\bigskip

Let $\Delta$ be the set of all probability distributions $\p=\{p_i:i\in\cal{N}\}$ on a countable
set $\cal{N}$: $p_i\ge 0$, $\sum_{i\in\cal{N}} p_i =1$.

In what follows, the inequalities between the vectors $\p>\q$ are understood component by
component: $p_i>q_i$ as $i\in\cal{N}$.

Let us expand the concept of relative entropy for infinite-dimensional probability vectors.
Let $\p=(p_i:i\in\cal{N})$, $\q=(q_i: i\in\cal{N})$ and $\q>\0$.

The relative entropy (Kullback-Leibler divergence) for the vectors $\p,\q\in\Delta$, $\q>\0$
is defined as
$$
D(\p\|\q)=\sum\limits_{i\in\cal{N}} p_i\ln\frac{p_i}{q_i}.
$$
We set $0\ln 0=0$.
Let us recall some properties of relative entropy that will be necessary in what follows~\citet{Vyu2022}.

\begin{lemma}\label{cor-1a}
1) For any $\p,\q,\w\in\Delta$, where $\q>{\bf 0}$ and $\w>\0$,
$$
D(\p\|\q)\le D(\p\|\w)+\ln\left(\sum\limits_{i\in\cal{N}} p_i\frac{w_i}{q_i}\right).
$$
2) If $\q\ge\beta\w$ for some number $\beta>0$, then
$$
D(\p\|\q)\le D(\p\|\w)+\ln\frac{1}{\beta}.
$$
3) In particular, $\p=\w$ and $\q\ge\beta\w$ will be $D(\w\|\q)\le\ln\frac{1}{\beta}$.
\end{lemma}

\emph{Proof}. From the concavity of the logarithm, we obtain inequality 1):
\begin{equation}\label{Vyugin 4.78}
D(\p\|\q)-D(\p\|\w)=\sum\limits_{i\in\cal{N}} p_i
\ln\frac{w_i}{q_i}\le\ln\sum\limits_{i\in\cal{N}} p_i \frac{w_i}{q_i}.
\end{equation}
If $\q\ge\beta\w$ then $D(\w\|\q)-D(\p\|\w)\le
\ln\sum\limits_{i\in\cal{N}} p_i\frac{w_i}{\beta w_i}\le\ln \frac{1}{\beta}$, i.e., 2) is satisfied.
Since $D(\p\|\w)=0$ for $\p=\w$, from 2) we get 3).
$\Box$

A mixing scheme (posterior distributions of experts) is a vector
${\bf \beta}=(\beta_0,\dots,\beta_t)$, where $\beta_i\ge 0$
for $0\le i\le t$ and $\sum_ {i=0}\beta_i=1$.

\begin{corollary}\label{cor-1}
Let ${\bf \beta}=(\beta_0,\ldots,\beta_t)$ be a mixing scheme,
$\w_s$ such that $\w_s>{\bf 0}$ for $0\le s\le t$.
Let also, $\q=\sum\limits_{s=0}^t\beta_i\w_s$ be a vector of a convex combination of vectors $\w_s$.
Then for an arbitrary vector $\p\in\Delta$ it will be
$$
D(\p\|\q)\le D(\p\|\w_s)+\ln\frac{1}{\beta_s}
$$
for any $s$ such that $\beta_s>0$.

In particular, for $\p=\w_s$ we have an estimate for the discrepancy between an arbitrary element $\w_s$
and vector of a convex combination:
$$
  D \left(\w_s\|\sum\limits_{i=0}^t \beta_i\w_i\right)\le\ln\frac{1}{\beta_s}.
$$
\end{corollary}

Here is a modified scheme for the weights update in Protocol 1 using the method of Mixing Past Posteriors -- MPP.

\bigskip

\begin{flushleft}
Parameter $\eta>0$.
We set $w_{i,1}=\tilde w_{i,0}=\frac{1}{c(i+1)\ln^2(i+1)}$ for $i\in\cal{N }$,
denote in vector form $\w_t=(w_{t,1},w_{t,2},\dots)$ and $\tilde\w_t=(\tilde w_{t,1},\tilde w_{t,2},\dots)$.

{\bf FOR} $t=1, \ldots, T$

Let at step $t$ experts incur their losses $l_{i,t}$ for $i\in{\cal N}$ and
Predictor incurs its loss $h_t$.

We update the expert weights in two stages:

{\bf Loss Update}
$$
\tilde w_{i,t} = \frac{w_{i,t}e^{-\eta l_{i,t}}}
{\sum\limits_{j\in\cal{N}}w_{j,t} e^{-\eta l_{j,t}}}
$$
for $i\in\cal{N}$.

{\bf Mixing Update}

Define the mixing scheme ${\bf \beta}^{t+1} = (\beta^{t+1}_0, \ldots, \beta^{t+1}_t)$ and update the weight of
the $i$th expert:
$$
w_{i,t+1} =\sum\limits_{s=0}^t \beta^{t+1}_s \tilde w_{i,s}
$$
$i\in\cal{N}$.

\textbf{ENDFOR}

\end{flushleft}

\bigskip

Below are examples of mixing schemes from \citet{BoW1998} and~\citet{Vyu2022}.

\textbf{Example 1.} $\beta_t^{t+1}=1$, where $\beta_s^{t+1}=0$, for $s=0, \ldots t$
(i.e., in convex combination weights in previous steps are not taken into account).
It turns out that the weights are corrected for the exponential scheme mixing (\ref{weight-1})
$$
w_{i,t+1}=\tilde w_{i,t}=\frac{w_{i,t} e^{-\eta l_{i,t}}}
{\sum\limits_{j\in\cal{N}} w_{j,t} e^{-\eta l_{j,t}}}
$$
from Protocol 1.

\textbf{Example 2.} $\beta_t^{t+1}=1-\alpha$, $\sum\limits_{s=0}^{t-1}\beta_s^{t+1}=\alpha$.
Any such scheme ${\bf\beta}^{t+1}$ is called Fixed-Share with the parameter $\alpha\in [0,1]$.
In particular, the following mixing scheme will be used: $\beta_0^{t+1}=\alpha$ and $\beta_s^{t+1}=0$ for $0<s<t$. For this mixing scheme
$$
w_{i,t+1}=\alpha\tilde w_{i,0} + (1-\alpha)\tilde w_{i,t}.
$$

Let $\l_t=(l_t^1, \ldots, )$ be the (infinite-dimensional) loss vector of all experts at the step $t$, $l_{i,t}\ge 0$
for all $i$ and $t$;
$m_t = -\frac{1}{\eta}\ln\sum\limits_{i\in\cal{N}}\w_{i,t} e^{-\eta l_{i,t}}$ -- exponentially mixed losses (mixloss)
at step $t$; $M_T = \sum\limits_{t=1}^T m_t$ -- cumulative mixloss over $T$ steps.

Denote $L_{i,T} = \sum\limits_{t=1}^T l_{i,t}$ -- cumulative loss of the expert $i$
$i, i=1, \ldots$; $H_T = \sum\limits_{t=1}^T h_t$ -- Predictor's cumulative loss.
for the first $T$ steps.

A vector $\q_t = (q_{i,t}: i\in\cal{N})$, where $\q_t\in\Delta$, is called a comparison vector if all its coordinates
are equal to 0 except for a finite number of them. Consider the convex combinations of expert losses
$(\q_t\cdot\l_t) = \sum\limits_{i\in\cal{N}} q_{i,t}l_{i,t}$ and weights $(\q_t \cdot\w_t) = \sum\limits_{i\in\cal{N}}q_{i,t}w_{i,t}$,
where $q_t = (q_{i,t}: i\in\cal{N})$ is the comparison vector.
A bound for the mixloss at step $t$ is given in the following theorem.

\begin{theorem}\label{cumul-1}
Let $\tilde\w_t=(\tilde w_{1,t},\dots)$ and $\w_t=(w_{1,t},\dots)$ be the weight vectors from the Loss Update and
Mixing update procedures.

For any $t$ and $0\le s<t$ such that $\beta_s^t>0$, and for any comparison vector $\q_t$,
\begin{eqnarray}
m_t\le (\q_t\cdot \l_t) + \frac{1}{\eta} (D(\q_t\|\w_t) - D(\q_t\|\tilde\w_t))\le
\label{Vyugin 4.79}
\\
\le (\q_t \cdot \l_t) + \frac{1}{\eta} \left( D(\q_t \|\tilde\w_s)-D(\q_t\|\tilde\w_t) + \ln \frac{ 1}{\beta_s^t}\right).
\label{Vyugin 4.80}
\end{eqnarray}
\end{theorem}

\emph{Proof:} Due to (\ref{Vyugin 4.78}),
\begin{eqnarray*}
m_t=- \frac{1}{\eta}\ln\sum\limits_{i\in{\cal N}} w_{i,t} e^{-\eta l_{i,t}}\le\sum\limits_{i\in\cal{N}} q_{i,t}
\left(-\frac{1}{\eta}\ln\sum\limits_{j\in\cal{N}} w_{j,t} e^{-\eta l_{j,t}} \right ) =
\\
=\sum\limits_{i\in\cal{N}} q_{i,t} \left( l_{i,t} + \frac{1}{\eta} \ln e^{-\eta l_{i ,t}} -
-\frac{1}{\eta} \ln \sum\limits_{j\in\cal{N}} w_{j,t} e^{-\eta l_{j,t}}\right)=
\\
=\sum\limits_{i\in\cal{N}} q_{i,t} l_{i,t} + \frac{1}{\eta} (D(\q_t\|\w_t)-D(\ q_t\|\tilde\w_t)).
\end{eqnarray*}

The inequality (\ref{Vyugin 4.80}) follows from (\ref{Vyugin 4.79}) by Corollary~\ref{cor-1}. $\Box$

\bigskip

Let's apply Theorem~\ref{cumul-1} for the mixing schemes from Examples 1 and 2.

\bigskip

\begin{corollary}\label{composite-expert-1}
For the mixing scheme ${\bf \beta}^{t+1}$ from Example 1, where $\beta_t^{t+1}=1$, and $\beta_s^{t+1}=0$ for all $0\le s<t$,
\begin{eqnarray}
M_T\le\sum\limits_{t=1}^T (\q\cdot\l_t)+\frac{1}{\eta} D(\q\|\w_1).
\end{eqnarray}
for any $T$ and for any comparison vector $\q$.
\end{corollary}

\emph{Proof.} Summing up the inequality (\ref{Vyugin 4.79}) with a constant comparison vector: $\q_t = \q$ for $t=1,\dots T$, we obtain
\begin{eqnarray}
  M_T\le\sum\limits_{t=1}^T (\q \cdot\l_t) + \frac{1}{\eta}\sum\limits_{t=1}^T (D(\q\| \w_t) - D(\q\| \tilde\w_t)) =
\nonumber
\\
\sum\limits_{t=1}^T (\q\cdot\l_t) +
\frac{1}{\eta} (D(\q \|\w_1)-D(\q\|\tilde\w_T))\le\sum\limits_{t=1}^T (\q \cdot \l_t) + \frac{1}{\eta}D(\q\|\w_1).
\label{summed-1}
\end{eqnarray}

Here, when passing from the first line to the second, we use the equality $\w_t = \tilde\w_{t-1}$,
which is the case for the mixing scheme from this example. Neighboring terms cancel and
only the first and last terms remain. Inequality (\ref{summed-1}) (satisfied,
since the first term satisfies $D(\q\|\tilde\w_T)\ge 0$. $\Box$

Let's estimate the losses for the mixing scheme from Example 2.

\begin{theorem}\label{composite expert-2}
Suppose that the comparison vector $q_t$ changes $k$ times
for $t=1, \ldots, T$: $k=|\{t: 1\le t\le T, \q_t\not=\q_{t- 1}\}|$.
Let $0<t_1<t_2< \ldots <t_k$ be the steps at which changes occur,
i.e. $\q_{t_j} \neq \q_{t_j-1}$ and $\q_t = \q_{t-1}$ for all other
steps $t$, $t>1$. We set $t_0=1$ and $t_{k+1}=T+1$.

For the mixing scheme from example 2, i.e. at
$\beta_t^{t+1}=1-\alpha$, $\beta_0^{t+1}=\alpha$, $\beta_s^{t+1}=0$, for $0<s<t$,
\begin{eqnarray}
M_T \le\sum\limits_{t=1}^T (\q_t\cdot \l_t)+\frac{1}{\eta}\sum\limits_{j=0}^k
\left(D(\q_{t_j}\|\w_1)-D(\q_{t_j} \|\tilde\w_{t_{j+1}-1}) \right) +
\nonumber
\\
\frac{1}{\eta} (k+1)\ln\frac{1}{\alpha}+\frac{1}{\eta} (T-k-1)\ln\frac{1}{1-\alpha}.
\label{Vyugin 4.81}
\end{eqnarray}
\end{theorem}

\emph{Proof.} Apply Theorem~\ref{cumul-1} to the distribution ${\bf \beta}^{t+1}$. Recall that
$\w_{i,1}=\tilde\w_{i,0}=\frac{1}{c(i+1)\ln^2(i+1)}$ for all $i$.
For any sequence $T$ of comparison vectors $\q_t$ with $k$ changes
\begin{eqnarray}
M_T\le\sum\limits_{t=1}^T (\q_t\cdot \l_t)+\frac{1}{\eta}
\sum\limits_{j=0}^k\left(D(\q_{t_j}\|\w_1)-D(\q_{t_j}\|\tilde\w_{t_{j+1}-1} )\right) +
\nonumber
\\
+\sum_{j=1}^k D(\q_{t_j}\|\tilde\w_0)+
\nonumber
\\
+\frac{1}{\eta} (k+1)\ln\frac{1}{\alpha}+\frac{1}{\eta} (T-k-1)\ln\frac{1}{1-\alpha}.
\label{Vyugin 4.81a}
\end{eqnarray}

Let us apply at each step $t$ the inequality (\ref{Vyugin 4.80}) from Theorem~\ref{cumul-1} for a suitable $s$:
For $t=1$ we put $s=0$, while $\beta_0^1=1$. We get
$$
m_1\le (\q_1\cdot\l_1) + \frac{1}{\eta} \left( D(\q_1\|\tilde\w_0)-D(\q_1\|\tilde\w_1)\right) .
$$
For those steps $t$, where the comparison vector did not change, i.e. $\q_t = \q_{t-1}$, we set $s=t-1$ and
use the property $\beta^t_{t-1} = 1-\alpha$ of the mixing scheme, i.e.,
$$
m_t\le (\q_t\cdot\l_t) + \frac{1}{\eta}\left(D(\q_t\|\tilde\w_{t-1})-D(\q_t\|\tilde\ w_t))\right)+ \frac{1}{\eta}\ln\frac{1}{1-\alpha}.
$$
For steps $t$, where the comparison vector was changed, $t=t_1, \ldots, t_k$,
we set $\beta_0^{t_j}=\alpha$ (for $s=0$), i.e.,
$$
m_{t_j}\le (\q_{t_j}\cdot\l_{t_j})+\frac{1}{\eta}\left(D(\q_{t_j}\|\tilde\w_0)-D( \q_{t_j}\|\tilde\w_{t_j}\right)+
+\frac{1}{\eta}\ln\frac{1}{\alpha}.
$$

We add up all these inequalities of three types. Terms of the same magnitude but different signs
inside the intervals will cancel, as in the proof of the Theorem~\ref{composite-expert-1}, for each partition interval,
only the initial points remain -- with a plus sign, and the end points -- with a minus sign,
these terms cancel each other out. In addition, the beginning of each interval corresponds
to an additional term $\frac{1}{\eta}\ln\frac{1}{\alpha}$, and each step $t$, where $\q_t = \q_{t-1 }$
corresponds to the additional term $\frac{1}{\eta}\ln\frac{1}{1-\alpha}$. There are only $k+1$ such additional terms
of the first type, and only $T-k-1$ of the second type. The sum $\sum_{j=1}^k D(\q_{t_j}\|\tilde\w_0)$ also remains. As a result,
we get (\ref{Vyugin 4.81a}).
$\Box$

Let the comparison vectors $\q_t$ have the form $\q_t=(0, \dots, 0, 1, 0,\dots )$, where the $i$-th coordinate is 1 and the rest
the coordinates are all equal to 0. In this case, at step $t$ we compare the loss of the algorithm with the loss
of only one $i$-th expert. In this case,
$D(\q_{t_j}\| \tilde\w_{0})=\ln (c(i+1)1\ln^2(i+1))\le\ln c + \ln (i+ 1)+2\ln\ln {i+1}$.\footnote{Recall that
that $c=\sum_{i=1}^T \frac{1}{\ln (i+1)\ln^2 (i+1}$.}

An arbitrary set $E$ of experts $i_0,i_1,\dots i_k$, and a set of intervals $[t_{j-1},t_j)$, $j=1,\dots k$. will be
called a composite expert, and its constituent experts will be called elementary. Since the total losses of the
elementary expert $i_j$ on the interval $[t_{j-1},t_j)$ are equal to $L_{([t_{j-1},t_j))}=\sum_{t_{j-1} \le t<t_j}l_{i_j,t}$,
the total losses of the composite Expert $E$ over the entire time interval $[1,T)$ are equal to $\sum_{j=1}^k L_{([t_{j-1},t_j))}$.

Let's set these losses with the help of comparison vectors. Consider a sequence of comparison
vectors $\q_1,\dots ,\q_T$ such that $\q_t=(0,\dots , 1, \dots, 0)$,
where the $i_j$-th coordinate is equal to 1 for $t_{ j-1}\le t<t_j$ and it is equal to 0, otherwise:
  \[
q_{i_j,t}=
\left\{
     \begin{array}{l}
       1, \mbox{ if } [t_{j-1}\le t<t_j),
     \\
       0 \mbox{ otherwise}.
     \end{array}
   \right.
\]
Then the total losses $L_T(E)$ of the composite expert $E$ on the entire interval $[0,T]$ can be represented as
$$
L_T(E)=\sum_{t=0}^T(\q_t\cdot\l_t)=\sum_{j=1}^k\sum_{t:t_{j-1}\le t<t_j}^ T q_{i_j,t}l_{j,t}=\sum_{j=1}^k L_{([t_{j-1},t_j)}.
$$
From Theorem~\ref{composite expert-2} we obtain an inequality relating the cumulative
exponentially mixed loss
and total losses of an arbitrary composite Expert.

\begin{corollary}
For any composite expert $E$ consisting of $k$ elementary experts, the inequality
\begin{eqnarray}
M_T \le L_T(E) + \frac{1}{\eta}(k+1)(\ln(T+1)+2\ln\ln(T+1)+\ln c)+
\nonumber
\\
\frac{1}{\eta}(k+1)\ln\frac{1}{\alpha}+\frac{1}{\eta} (T-k-1)\ln\frac{1}{1- \alpha}.
\label{regret-with-1}
\end{eqnarray}
\end{corollary}
\emph{Proof.} We will use the bound~\ref{Vyugin 4.81}.
Since $\sum_{j=1}^k D(\q_{t_j}\|\tilde\w_0)\le \ln\frac{1}{w_0}\le\ln(i_j+1)+2\ ln\ln(i_j+1)+\ln c$,
we get $\sum_{j=1}^k D(\q_{t_j}\|\tilde\w_0)\le k(\ln(i_j+1) +2\ln\ln(i_j+1)+\ln c)$. From here and
from~\ref{Vyugin 4.81} we get the bound~\ref{regret-with-1}.
$\Box$
\subsection{Aggregating Algorithm {\bf AA}}\label{AA}

The Aggregating Algorithm ({\bf AA}) proposed in~\citet{Vov1990},~\citet{Vov1998} is the basic
method for calculating Predictor predictions in this work.
Let ${\bf f}=(f_1,f_2,\dots)$ be the forecasts of $i\in\cal{N}$ experts and $\p=(p_i:i\in\cal{N})$ --
probability distribution on the set ${\cal N}$ of all experts.\footnote{That is, $p_i\ge 0$ for all $i$ and $\sum\limits_{i\in\cal{ N}}p_i=1$.}
The superprediction function is defined as
\begin{eqnarray*}
g(y)=-\frac{1}{\eta}\ln\sum\limits_{i\in\cal{N}} e^{-\eta\lambda(f_i,y)}p_i
\label{superpred-1}
\end{eqnarray*}
for arbitrary $y$, where $\eta>0$ is the learning rate~\citet{Vov1998}.\footnote{The series on the right side of (\ref{superpred-1})
converges, since the loss function takes non-negative values. }

A loss function $\lambda$ is said to be $\eta$-mixable if for any probability distribution $\p$ on
a set of experts and for any set of expert predictions ${\bf f}$ there exists a prediction $\gamma\in\Gamma$
which satisfies the inequality
\begin{equation}
\lambda(\gamma,y)\le g(y)
\label{subst-1}
\end{equation}
for all $y$.

We fix some rule $\gamma=\subst({\bf f},\p)$ for computing the prediction $\gamma$, satisfying (\ref{subst-1}).

%\footnote{If such a rule can be defined.}
The $\subst$ function is called the substitution function.

In what follows, we will use the square loss function $\lambda(\gamma,y)=(y-\gamma)^2$,
where $y$ $\gamma$ are real numbers. we assume that $y\in [a,b]$, where $a<b$ are real
numbers

In~\citet{Vov1998} and~\citet{Vov2001} it is proved that the square loss function is $\eta$-mixable
for every $\eta$ such that $0<\eta\le\frac{2}{(b-a )^2}$, and the corresponding prediction is
\begin{eqnarray}
\gamma=\subst({\bf f},\p)=\frac{a+b}{2}+ \frac {1}{2\eta (b-a)}\ln\frac{\sum\limits_ {i\in \cal {N}} p_i e^{-\eta (b-f_i)^2}}
{\sum\limits_{i\in\cal{N}} p_i e^{-\eta (a-f_i)^2}}.
\label{su-1}
\end{eqnarray}
% The inequality (\ref{subst-1}) is also true for all $y$.
\begin{theorem} (\citealt{Vov1998} and~\citealt{Vyu2022})
Suppose that the loss function $\lambda(f,y)$ is $\eta$-mixable for some $\eta>0$.
Let $H_T$ be the total losses of the Predictor, and $L_{i,T}$ be the total losses of the Expert $i$.
Then for each $T$ the inequality
$H_T\le M_T\le L_{i,T}+\frac{1}{\eta}\ln\frac{1}{w_{i,1}}$.
\end{theorem}
\emph{Proof}. According to (\ref{subst-1})
\begin{eqnarray}
h_t=\lambda(f_t,y_t)\le g_t(y_t)=m_t
\label{h-t-bound-1-1}
\end{eqnarray}
for every $t$.
We sum the inequalities (\ref{h-t-bound-1-1}) over $t=1\dots ,T$ and get $H_T\le M_T$. Hence, by Proposition~\ref{prop-1s} for any $i$ and all $T$
done %\begin{eqnarray*}
$H_T\le L_{i,T}+\frac{1}{\eta}\ln\frac{1}{w_{i,1}}$.
%\end{eqnarray*}
$\Box$

\section{Algorithm for tracking of subsample generators}\label{main-1}

In this section, we present a prediction algorithm -- {\bf GMPP}.

Let us first motivate the method underlying the algorithm.

The general scheme of the online learning process is as follows. At each step $t$ one observes
signal $\x_t$. Expert strategies built on steps $1,2,\dots ,t$ present their response predictions. For simplicity,
 we assume that $\x_t\in{\cal R}^n$, $y_t\in\cal{R}$. The predictor also presents his prediction. After that, the corresponding
generator $G$ produces the true response $y_t=G(\x_t)$, and the experts and the Predictor calculate their losses due to
the difference in their predictions and response.

There are $k+1$ generators that transform the signal $\x_t$ into the response $y_t$. The time interval $[0,T]$ is divided into subintervals,
on each of which one of these generators produces responses. At each time $t$, neither the experts nor Predictor know the number of
generators, and also which of the generators produces a response.

The described generation model creates a sample
$(\x_1,y_1),(\x_2,y_2),\dots$ which is divided into subsamples,
the responses $y_t$ in which are obtained as a result of the operation of one of the generators.

We assume that there is a learning method with the help of which at any time $t$
by subsample (window to the past), you can build a local predictive model
(expert).\footnote{A window into the past at time $t$ is understood as a subsample
$(\x_{t-1},y_{t-1},\dots , \x_{t-h},y_{t-h})$, where $h>0$ is a parameter (window size).
As such a method, the ridge regression method will be used.
In section~\ref{local-exp-1}, the prognostic model (Expert)
will be given by the regression equation $y=(\a\cdot\x)$, where $\a\in {\cal R}^n$.}

At each step $t>h$, a prognostic model is built (initialized) -- a function $f_t(\x)$,
which is determined by the previous observed members of the time series --
by a window into the past
$$
(\x_{t-1},y_{t-1},\dots , \x_{t-h},y_{t-h}),
$$
where $h$ is a parameter (window size).

Thus, at each step $t$ there is a collection of predictive strategies (models)
$\f_t=(f_1,\dots,f_{t-1})$ constructed at the previous steps and the predictive function $f_t$ constructed at the step $t$.

Expert's $i\le t$ forecast at step $t$ is equal to $f_{i,t}=f_i(\x_t)$, where $\x_t$ is the signal at step $t$,

Calculate the forecast $\gamma_t$ Predictor according to the rule (\ref{su-1}).
At the steps $t<i$, when $i$ has not yet been initialized, we introduce a virtual
forecast -- we will assume that his forecast is equal to the forecast $\gamma_t$
of the predictor (aggregation algorithm).

This definition contains a logical circle since the prediction $\gamma_t$ of the Predictor is defined by
aggregation of forecasts of all experts, including experts $i>t$. This contradiction will be resolved using
the fixed point method proposed in~\citet{ChV2009} as follows.
Let's assume that Predictor's $\gamma_t$ forecast is known to experts.
Define forecasts of the experts $i=1,2,\dots$ at step  $t$:
\[
  f_{i,t}
  =
  \left\{
    \begin{array}{l}
      f_i(\x_t), \mbox{ if } i\le t,
    \\
      \gamma_t, \mbox{ if } i>t.
    \end{array}
  \right.
\]
The loss of the aggregation algorithm is $h_t=\lambda(\gamma_t,y_t)$, and the loss of any expert $i$ is
$l_{i,t}=\lambda(f_{i,t},y_t)$ for $i\le t$ and $l_{i,t}=h_t$ for $i>t$.
 forecast $\gamma_t$ should satisfy the condition
\begin{eqnarray}
\lambda(\gamma_t,y))\le g_t(y),
\label{for-1b-0}
\end{eqnarray}
 or, equivalently, the condition
\begin{eqnarray}
e^{-\eta\lambda(\gamma_t,y)}\ge\sum_{i\in\cal{N}}
e^{-\eta\lambda(f_{i,t},y)}w_{i,t}
\label{for-1b-1}
\end{eqnarray}
have to satisfy for every $y$.

Let's replace the condition (\ref{for-1b-1}) with an equivalent condition under which the summation is performed over a finite set of experts.
Since $f_{i,t}=f_i(\x_t)$ for $i\le t$ and $f_{i,t}=\gamma_t$
for $i>t$, we present the condition (\ref{for-1b-1}) for $\gamma_t$ in a more detailed form:
\begin{eqnarray}
e^{-\eta\lambda(\gamma_t,y)}\ge
\sum_{i=1}^t w_{i,t}e^{-\eta\lambda(f_i,y)}+
e^{-\eta\lambda(\gamma_t,y)}\left(1-\sum_{i=1}^t w_{i,t}\right).
\label{cond-2}
\end{eqnarray}
Thus, the inequality (\ref{for-1b-1}) is equivalent to the inequality
\begin{eqnarray}
e^{-\eta\lambda(\gamma_t,y)}\ge\sum_{i=1}^t
w^p_{i,t}e^{-\eta\lambda(f_{i,t},y)},
\label{for-1ba}
\end{eqnarray}
Where
\begin{eqnarray}\label{for-1bb}
w^p_{i,t}=\frac{w_{i,t}}{\sum_{j=1}^t w_{j,t}}.
\end{eqnarray}
According to the rule (\ref{su-1}) for {\bf AA}, we define
\begin{equation}\label{subst-2}
\gamma_t=\subst({\bf f}_t,\w^p_t),
\end{equation}
where $\subst$ is the substitution function for the loss function used,\footnote{For example,
for a quadratic loss function, the substitution function is defined according to (\ref{su-1}).}
where $\w^p_t=(w^p_ {1,t},\dots ,w^p_{t,t})$ and
${\bf f}_t=(f_1 (\x_t),\dots ,f_t(\x_t))$.

From the definition for $y=y_t$ it will be
$$
h_t=\lambda(\gamma_t,y_t))\le g_t(y_t)=m_t,
$$
where $m_t$ is the exponentially mixed loss. We summarize this inequality
over $t=1,\dots ,T$ and get $H_T\le M_T$.

The expert weights are are updated in two stages as follows:

  {\bf Loss Update}

  \begin{equation}\label{loss-update-1}
    \tilde w_{i,t} = \frac{
w_{i,t}e^{-\eta l_{i,t}}
}
{
\sum\limits_{j\in\cal{N}}w_{j,t} e^{-\eta l_{j,t}}
}
  \end{equation}
for $i\in\cal{N}$.

  {\bf Mixing Update}

\begin{equation}\label{mixing-update-1}
w_{i,t+1}=\alpha_t \tilde w_{i,1}+(1-\alpha_t) \tilde w_{i,t}
\end{equation}
for $i\in\cal{N}$, where $\alpha_t$ is a parameter, $0<\alpha_t<1$.

From the definition (\ref{loss-update-1}) it follows that
$$
\sum_{j\in\cal{N}}
w_{j,t}=1
$$
for each $t$. It follows from (\ref{mixing-update-1}) that
$$
\sum_{j\in\cal{N}}
\tilde w_{j,t}=1
$$
for any $t$.

Recall that the Predictor's loss is $h_t=\lambda(\gamma_t,y_t)$, and the experts losses are
are $l_{i,t}=\lambda(f_{i,t},y_t)$ for $i\le t$ and $l_{i,t}=h_t$ for $i>t$. Using these equalities, we represent the sum

in the denominator  of (\ref{loss-update-1}) in a computationally efficient form
\begin{eqnarray*}
\sum\limits_{j\in\cal{N}}w_{j,t} e^{-\eta l_{j,t}}=
\nonumber
\\
\sum\limits_{j\le t}w_{j,t}e^{-\eta l_{j,t}}+
\sum\limits_{j>t}w_{j,t} e^{-\eta h_t}=
\nonumber
\\
\sum\limits_{j\le t}w_{j,t}e^{-\eta l_{j,t}}+
e^{-\eta h_t}\sum\limits_{j>t}w_{j,t} =
\nonumber
\\
\sum\limits_{j\le t}w_{j,t}e^{-\eta l_{j,t}}+
e^{-\eta h_t}(1-\sum\limits_{j\le t}w_{j,t}) =
\nonumber
\\
\sum\limits_{j\le t}w_{j,t}e^{-\eta l_{j,t}}+
e^{-\eta h_t}(1-\sum\limits_{j\le t}w_{j,t}).
\end{eqnarray*}
Therefore, the (\ref{loss-update-1}) {\bf Loss Update} part is replaced with the following definition:
  \begin{eqnarray}
\tilde w_{i,t}=\frac{
w_{i,t}e^{-\eta l_{i,t}}}
{
\sum\limits_{j=1}^t w_{j,t} e^{-\eta l_{j,t}}+
e^{-\eta h_t}(1-\sum\limits_{j=1}^t w_{j,t})
},
\label{loss-update-2}
\end{eqnarray}
and the {\bf Mixing update} part is still (\ref{mixing-update-1}).

Let's present the protocol of the {\bf GMPP} algorithm.
Let's first set the parameters $\eta$ and $\alpha_t$, where $0<\alpha_t<1$ for $t=1,2\dots$, and $\eta>0$.%
\footnote{For the square loss function, we set $\eta=\frac{2}{(b-a)^2}$, where $y_t\in[a,b]$.}
We put $\alpha_t=\frac{1}{t+1}$ for all $t$.

\medskip

\noindent{\bf Algorithm {\bf GMPP}}
{\small
\medskip\hrule\medskip

\noindent Define the initial weights $w_{i,1}=\tilde w_{i,0}$ of the experts such that
$\sum_{i\in\cal{N}}w_{i,1}=1$.\footnote{For example,
$w_{i,1}=\tilde w_{i,0}=\frac{1}{c(i+1)\ln^2(i+1)}$ for $i=1,2,\dots$,
where $c=\sum_{i\in\cal{N}}\frac{1}{(i+1)\ln^2(i+1)}$, $\frac{1}{\ln 3}<c<\frac{1}{\ln 2}$.}

\noindent\hspace{2mm}{\bf FOR} $t=1,\dots ,T$

\begin{enumerate}
\item
Experts $f_1(\cdot),\dots ,f_{t-1}(\cdot)$
have been initialized in the previous steps.

%%\noindent\hspace{2mm}{\bf IF} $t\le h$

Initialize the Expert $f_t(\cdot)$.\footnote{In the case when the regression problem is
being solved, initialization means that
we use the data from the past to determine the weight vector $\a_t$ of the regression equation
$f_t(\x)=(\a_t \cdot\x)$.}

%\noindent\hspace{2mm}{\bf ElSE}
%\smallskip

\setcounter{enumi}{2}
\item
We receive the signal $\x_t$.
\item
Calculate expert forecasts $f_{i,t}=f_i(\x_t)$ for $1\le i\le t$.
\item
Calculate auxiliary weights of the experts$1\le i\le t$:
\begin{eqnarray}
w^p_{i,t}=\frac{w_{i,t}}{\sum_{j=1}^t w_{j,t}}.
\label{w-u-3}
\end{eqnarray}
\item
Calculate the Predictor's forecast according to the rule (\ref{su-1}):
$$
\gamma_t=\subst({\bf f}_t,\w^p_t),
$$
\item
We get (from the generator) the true value of the sign (label) $y_t$ and calculate
the loss $h_t=\lambda(\gamma_t,y_t)$ of Predictor's and the experts losses:
\[
l_{i,t}=
\left\{
     \begin{array}{l}
       \lambda(f_{i,t}, y_t)\mbox{ if } i\le t,
     \\
       h_t \mbox{ if } i>t.
     \end{array}
   \right.
\]

\item
We update the weights of the experts $1\le i\le T$ in two stages:\footnote{Thus,
it is assumed that the prediction horizon $T$ is given to Predictor as a parameter.}

    {\bf Loss Update}

%for $i\le t+1$
\begin{equation}\label{loss-update-2a}
\tilde w_{i,t}=\frac{
w_{i,t}e^{-\eta l_{i,t}}
}
{
\sum\limits_{j=1}^t w_{j,t} e^{-\eta l_{j,t}}+
e^{-\eta h_t}(1-\sum\limits_{j=1}^t w_{j,t})
}.
\end{equation}

{\bf Mixing Update}

\begin{equation}\label{mixing-update-2a}
w_{i,t+1}=\alpha_t \tilde w_{i,1}+(1-\alpha_t)\tilde w_{i,t}
\end{equation}

\end{enumerate}

{\bf ENDFOR}

\medskip\hrule\medskip
A bound of the efficiency of the {\bf GMPP} algorithm is presented in the following theorem.

\begin{theorem} \label{main-theorem-1}
Let $\alpha_t=\frac{1}{(t+1)}$ for all $t$. For any composite expert $E$ consisting of $k+1$ elementary experts,
there will be
\begin{eqnarray}
%M_T\le\sum\limits_{t=0}^T (\q_t\cdot \l_t) + \frac{1}{\eta}
%\sum\limits_{j=0}
M_T\le L_T(E)+\sum_{j=1}^k D(\q_{t_j}\|\tilde\w_0)+\frac{1}{\eta} (k+1) \ln c +
\nonumber
\\
\frac{1}{\eta} (k+1)
(\ln(T+1)+2\ln\ln(T+1)+\ln c+
\ln T) + \frac{1}{\eta}\ln (T-k-1)
\label{regret-with-2}
\end{eqnarray}
for all $T$, where $L_T(E)$ is the total loss of the composite Expert. In addition, $H_T\le M_T$.
From the bound (\ref{regret-with-2}) it follows that
$$
\limsup_{T\infty}\frac{1}{T}(H_T-L_T(E))=0.
$$
\end{theorem}
\emph{Proof.}
Let's refine the process of obtaining the bound (\ref{regret-with-1}) of Corollary~\ref{composite-expert-1}                                                        a`
in the case when $\alpha_t=\frac{1}{t+1}$.
The score of the regret consists of three sums.

The first sum is
$\sum_{j=1}^k D(\q_{t_j}\|\tilde\w_0)$.
 $i_j\le T$ for all $j$ and $\q_{t_j}$ is the unit vector,
we have $D(\q_{t_j}\|\tilde\w_0)\le\ln T+2 \ln\ln T+\ln c$, so the first sum is bounded by
$\frac{1}{\eta}(k+1)(\ln(T+1)+2\ln\ln(T+1)+\ln c)$.

The second sum is
$$
\frac{1}{\eta}\sum_{t=1}^{k+1}\ln\frac{1}{t+1}.
$$
The third sum is
$$
\frac{1}{\eta}\sum_{t=1}^{T-k}\ln\frac{1}{1-\frac{1}{t+1}}=\frac{1}{\eta }\sum_{t=1}^{T-k-1}((\ln(t+1)-\ln t).
$$
We restrict the second sum to $\frac{1}{\eta}(k+1)\ln T$. We restrict the third sum
to $\frac{1}{\eta}\ln (T-k-1)$.
Using these considerations and the bound (\ref{regret-with-1}), we obtain the bound (\ref{regret-with-2}).
$\Box$

Let the intervals $[t_1,t_2), \dots, (t_{j-1},t_j], (t_{k-1}, t_k]$ define the data areas generated by
corresponding generators.
Let us introduce a composite Expert $E$ consisting of elementary experts $i_1,\dots ,i_k$ and
the corresponding intervals
$$
(t_1,t_2], \dots, (t_{j-1},t_j], (t_{k-1}, t_k],
$$
where for each $j\le k$ the elementary expert $i_j$, which was initialized on some interval
$\le t_j$ and bears
the least loss on the interval $[t_{j-1}, t_j)$
among all experts initialized at steps $\le t_j$, By Theorem~\ref{main-theorem-1}, the
bound (\ref{regret-with-2}) takes place.

The bound (\ref{regret-with-2}) allows us to formulate the main
hypothesis underlying the application of the {\bf GMPP} algorithm:
In the case where it is possible to ``attach'' to each local subsample from the generation area
valid predictive (expert) strategy,
carrying low loss on each local subsample generated
by the generator, i.e., ``learn'' ''this generator,
the {\bf DMPP} algorithm will also predict with sufficiently small average (in time)
loss over the entire sample.

\subsection{Numerical experiments}\label{local-exp-1}

Time scale $[1,T]$ is divided into $k=10$ consecutive time intervals $I_1,\dots,I_k$,
on which one  each of fours generators are performed.
% a regression function).
Therefore,
on the time interval $[1,T]$ the dependence of $y_t$ on $\x_t$
is switched $k=9$ times.

The number of generators and their parameters are unknown to Forecaster.
$e=4$ linear response generators are used, defined by weight vectors $\hat\a_1,\dots,\hat\a_e$,
i.e. within the corresponding generation interval $I_s$ the response is equal to
$y_t=(\hat\a_s\cdot \x_t)+\epsilon$ for $1\le s\le e$, where $\epsilon$ is the standard normal noise.

% $((\x_{i-h},y_{t-h}),\dots ,(\x_{i-1},y_{i-1}))$.
% Используется квадратичная функция потерь $\lambda(\gamma,y)=(\gamma-y)^2$.

At each step $t$ using the ridge regression method over a window into the past
$((\x_{i-h},y_{t-h}),\dots ,(\x_{i-1},y_{i-1})$
an expert predictive function is constructed $f_t(\x)=(\a_t\cdot\x)$.\footnote
{Here $$
\a_t=\left(\sigma I+X'_t X_t\right)^{-1} X_t' \y_t
$$
for $t>h$.
Here $X_t$ is a matrix whose columns are formed by the vectors
$\x_{t-h},\dots , \x_{t-1}\in {R}^n$, where $X'_t$ is the transposed matrix
$X_t$, $I$ -- identity matrix, $\sigma>0$ is a parameter, $h$ is the window size and
$\y_t=(y_{t-h},\dots , y_{t-1})$.
For $t\le h$, we take some fixed vector as $\a_t$.
}

$$
\f_t=(f_1,\dots ,f_t),  \x_{t-h},y_{t-h}.
$$

%On Fig.~\ref{fig-1a} the  curves of the time averages of the total losses of the algorithm GMPP
%and the same curve of for the valid composite expert
%\begin{figure}[htb]
%  \centering
%    \includegraphics[width=0.999\linewidth]{figure1.png} 
%\caption{Curves of time averages of total losses of the algorithm GMPP (upper curve)
%  and the same for composite expert}\label{fig-1a}
%\end{figure}

\section{Conclusion }

An online learning algorithm is presented for tracking online generators of subsamples.

An evident drawback of the computation scheme is that on each step $1\le t\le T$ we perform  computationally expensive operations
(\ref{loss-update-2}) and (\ref{mixing-update-2a}) for each expert $1\le i\le T$,
and store the corresponding weights; in practical applications, restrictions on the number of
initialized experts were used.

\end{document}